\documentclass[preprint,12pt]{elsarticle}

\usepackage{amssymb}
\usepackage{amsmath}
\usepackage{booktabs,tabularx,adjustbox,siunitx, subcaption}
\usepackage{array}
\usepackage{xcolor}
\usepackage{multirow}
\journal{Image and Vision Computing}

\begin{document}

\begin{frontmatter}

%% Title, authors and addresses

%% use the tnoteref command within \title for footnotes;
%% use the tnotetext command for theassociated footnote;
%% use the fnref command within \author or \affiliation for footnotes;
%% use the fntext command for theassociated footnote;
%% use the corref command within \author for corresponding author footnotes;
%% use the cortext command for theassociated footnote;
%% use the ead command for the email address,
%% and the form \ead[url] for the home page:
%% \title{Title\tnoteref{label1}}
%% \tnotetext[label1]{}
%% \author{Name\corref{cor1}\fnref{label2}}
%% \ead{email address}
%% \ead[url]{home page}
%% \fntext[label2]{}
%% \cortext[cor1]{}
%% \affiliation{organization={},
%%             addressline={},
%%             city={},
%%             postcode={},
%%             state={},
%%             country={}}
%% \fntext[label3]{}

\title{Beyond the Generative Learning Trilemma: Generative Model Assessment in Data Scarcity Domains} %% Article title

%% use optional labels to link authors explicitly to addresses:
%% \author[label1,label2]{}
%% \affiliation[label1]{organization={},
%%             addressline={},
%%             city={},
%%             postcode={},
%%             state={},
%%             country={}}
%%
%% \affiliation[label2]{organization={},
%%             addressline={},
%%             city={},
%%             postcode={},
%%             state={},
%%             country={}}

\author[aff1]{Marco Salmè}

\author[aff1]{Lorenzo Tronchin}

\author[aff1]{Rosa Sicilia}

\author[aff1,aff2]{Paolo Soda}\corref{cor}
\ead{p.soda@unicampus.it, paolo.soda@umu.se}

\author[aff1]{Valerio Guarrasi}

%% Author affiliation

\affiliation[aff1]{organization={Unit of Computer Systems and Bioinformatics, Department of Engineering \\ Università Campus Bio-Medico di Roma},
            city={Rome},
            country={Italy}}

\affiliation[aff2]{organization={Department of Diagnostics and Intervention, Radiation Physics, Biomedical Engineering \\ Umeå University},
            city={Umeå},
            country={Sweden}}

%% Abstract
\begin{abstract}
Data scarcity remains a critical bottleneck impeding technological advancements across various domains, including but not limited to medicine and precision agriculture. To address this challenge, we explore the potential of Deep Generative Models (DGMs) in producing synthetic data that satisfies the Generative Learning Trilemma: fidelity, diversity, and sampling efficiency.  However, recognizing that these criteria alone are insufficient for practical applications, we extend the trilemma to include utility, robustness, and privacy, factors crucial for ensuring the applicability of DGMs in real-world scenarios. Evaluating these metrics becomes particularly challenging in data-scarce environments, as DGMs traditionally rely on large datasets to perform optimally. This limitation is especially pronounced in domains like medicine and precision agriculture, where ensuring acceptable model performance under data constraints is vital. To address these challenges, we assess the Generative Learning Trilemma in data-scarcity settings using state-of-the-art evaluation metrics, comparing three prominent DGMs: Variational Autoencoders (VAEs), Generative Adversarial Networks (GANs), and Diffusion Models (DMs). Furthermore, we propose a comprehensive framework to assess utility, robustness, and privacy in synthetic data generated by DGMs. Our findings demonstrate varying strengths among DGMs, with each model exhibiting unique advantages based on the application context. This study broadens the scope of the Generative Learning Trilemma, aligning it with real-world demands and providing actionable guidance for selecting DGMs tailored to specific applications.

\end{abstract}

%%Graphical abstract
%\begin{graphicalabstract}
%\includegraphics{grabs}
%\end{graphicalabstract}

%%Research highlights
%\begin{highlights}
%\item Research highlight 1
%\item Research highlight 2
%\end{highlights}

%% Keywords
\begin{keyword}
%% keywords here, in the form: keyword \sep keyword
 Deep Generative Models \sep Data Scarcity \sep Synthetic Data \sep Medicine \sep Precision Agriculture \sep Generative Learning Trilemma
%% PACS codes here, in the form: \PACS code \sep code

%% MSC codes here, in the form: \MSC code \sep code
%% or \MSC[2008] code \sep code (2000 is the default)

\end{keyword}

\end{frontmatter}

%% Add \usepackage{lineno} before \begin{document} and uncomment 
%% following line to enable line numbers
%% \linenumbers

%% main text
%%

%% Use \section commands to start a section
\section{Introduction}
\label{intro}

In the digital era, data scarcity poses a significant challenge to the development of innovative solutions across various application domains~\cite{bansal2022systematic}. Precision agriculture, for instance, is hindered by fluctuating weather conditions and terrain variations~\cite{karunathilake2023path}. Similarly, in medicine, data scarcity can arise from factors such as the limited number of patients for certain diseases or the lack of standardized protocols~\cite{ibrahim2021health}. Additionally, the medical field faces constraints imposed by privacy regulations, which necessitate the complete anonymization of data in the absence of patient consent~\cite{olatunji2022review}. Data scarcity is also one of the main causes that can jeopardize the achievement of effective training training of Deep Learning (DL) models, as it can lead to overfitting~\cite{ying2019overview}. Data Augmentation (DA) techniques, which are based on the addition of synthetic data during the training of DL models, can be adopted to mitigate this problem. The most popular DA technique in the literature applies affine transformations to the original images, such as rotations, translations, scaling, and shearing~\cite{chlap2021review}.  
However, this approach requires knowledge of the application domain: for example, in the context of medical images, it is essential that these transformations respect anatomical features. Similarly, in the context of precision agriculture, transformations must preserve the spatial patterns of crops, soil textures, and vegetation indices~\cite{radovcaj2023state}. 
An alternative approach involves the use of Deep Generative Models (DGMs), which have demonstrated remarkable success in producing realistic and varied synthetic data that exhibit high fidelity and diversity~\cite{eigenschink2023deep}. 

Current generative learning frameworks face significant challenges in simultaneously meeting three essential requirements, which hinders their widespread adoption in real-world applications.
These requirements, commonly referred to as the Generative Learning Trilemma~\cite{xiao2021tackling}, encompass fidelity, diversity, and sampling speed. 
Specifically, the first requirement, \textit{fidelity}, refers to ensuring that the generated
data are both realistic and high-quality. This means they must accurately replicate
the fine-grained details and distinctive attributes of the target data distribution.
The second requirement, \textit{diversity}, ensures that the generated data are varied. The third requirement, \textit{sampling speed}, refers to the rapid generation of synthetic data by a DGM.
The main categories of generative learning frameworks include Generative Adversarial Networks (GANs), Variational Autoencoders (VAEs), and Diffusion Models (DMs). GANs~\cite{goodfellow2014generative} are capable of producing high-quality samples rapidly; however, they often struggle with data diversity~\cite{salimans2016improved}. In contrast, VAEs~\cite{kingma2013auto} excel at accurately covering data modes with rapid sampling generation, yet they frequently produce lower-quality samples. Recently, a novel class of generative models known as DMs~\cite{denoisingDiffusionProbModels} has emerged, which have demonstrated exceptional performance in image generation, surpassing the quality achieved by GANs~\cite{dhariwal2021diffusion}. Additionally, DMs show a robust capacity for generating diverse images~\cite{song2021maximum}, however, their primary limitation is their sampling speed, which significantly restricts their practicality in real-world scenarios.

Although the requirements introduced in the trilemma are essential, they remain insufficient for real-world applications, as additional challenges beyond the three encompassed in the trilemma must also be considered. 
Therefore, we propose extending the generative learning trilemma by
incorporating metrics that address increasingly critical aspects, such as \textit{utility}, \textit{robustness} and \textit{privacy}. 
The term \textit{utility} refers to the effectiveness of a DGM in practical applications: it measures how well a DGM supports performance in downstream tasks, such as classification, segmentation, or regression. 
For instance, consider a precision agriculture system that utilizes synthetic data generated by a DGM to predict crop growth under varying climatic conditions. 
If the synthetic data fails to accurately capture the specific needs of the plants or the characteristics of the soil, the models may make suboptimal decisions, such as over-irrigation or inefficient fertilisation. 
Assessing the utility of the generated data is essential, as only data that improves predictions and operational decision-making can be considered effective. Otherwise, the use of such data may degrade model performance and lead to detrimental outcomes. Given the increasing risk of adversarial attacks~\cite{chakraborty2021survey}, it is also crucial to assess the \textit{robustness} of the synthetic data generated by these models. 
By investigating whether using synthetic data for DA can enhance the robustness of DL models to these attacks. Evaluating robustness is particularly crucial in the medical domain because adversarial attacks on DL models could lead to misdiagnosis, inappropriate treatments, and compromised patient safety by imperceptibly
altering medical imaging~\cite{rodriguez2022role}. 
Additionally, DL models for medical imaging are more vulnerable to adversarial attacks than those used for natural images, as successful attacks on medical images can be achieved with smaller perturbations~\cite{ma2021understanding}.
Moreover, while DGMs offer a promising approach to generating fully anonymized data for practical use, further evaluation is required to determine which DGM provides the highest level of \textit{privacy} protection. 
A significant concern with DGMs is the risk of data memorization, wherein a model inadvertently memorizes and reproduces training data, thus compromising privacy. This risk is particularly dangerous in sensitive fields such as medical applications, where DGMs could address data scarcity issues but might simultaneously violate privacy standards, rendering them unsuitable for such domains.

This paper introduces the following key contributions to the field of synthetic data generation using DGMs:
\begin{itemize}
    \item We expand the conventional Generative Learning Trilemma, focusing on fidelity, diversity, and sampling speed, by incorporating three additional metrics essential for practical applications: utility, robustness, and privacy. 
    \item We introduce a robust experimental framework to evaluate DGMs, i.e. GANs, VAEs, and DMs, across the extended set of six criteria, using four public datasets from medicine and precision agricolture, which are two scenarios marked by data limitations.
    \item We analyze the impact of data scarcity on DGM performance, assessing how different DGMs cope with limited training data and identifying strategies to leverage their generated data to enhance the training of other models.
    \item Through systematic testing in data-scarce environments, our study provides nuanced insights into the strengths and limitations of each DGM type. 
    The findings offer actionable guidance for practitioners on selecting appropriate DGMs tailored to specific application needs. 
\end{itemize}
By addressing these points, our research contributes substantially to both the theoretical and practical aspects of synthetic data generation, helping mitigate the impact of data scarcity across diverse application domains.

\section{State-of-the-art}
\label{sota}

The three requirements introduced in the Generative Learning Trilemma are frequently addressed in various studies~\cite{dhariwal2021diffusion, bayat2023study,yang2022your,jing2022subspace}. However, there is currently a lack of research that compares the main families of DGMs (VAEs, GANs, and DMs) with respect to these requirements simultaneously, namely fidelity, diversity, and sampling speed.
Studies have focused almost exclusively on comparing GANs and DMs, primarily due to the limitations demonstrated by VAEs in terms of generation quality and scalability to larger image sizes~\cite{bond2021deep}.
Dhariwal et al.~\cite{dhariwal2021diffusion} have shown that DMs are able to generate images with higher quality than GANs, while also offering more diversity. 
The superiority of DMs over GANs in terms of diversity is also confirmed in other studies~\cite{bayat2023study}.
However, despite continuous advancements in computational efficiency, DMs remain significantly slower than GANs in generating samples~\cite{yang2022your, jing2022subspace}.

To ensure synthetic data are meaningful, they must strike a balance by being both similar to and different from the original data. 
This involves the addition of three other fundamental requirements: utility, privacy and robustness, already defined in section~\ref{intro}. 
Much of the existing literature on synthetic data, emphasizes the two-dimensional trade-off between utility and privacy, often incorporating fidelity within the concept of utility~\cite{ullman2011pcps}. 
While utility and fidelity are inevitably interconnected, they are neither synonymous nor perfectly correlated. 
The utility of a synthetic dataset is entirely dependent on its intended application. The most prevalent method for assessing utility is the \textit{Train on Synthetic, Test on Real} paradigm~\cite{esteban2017real}, in which models are trained on synthetic data and their performance is subsequently evaluated on real data~\cite{beaulieu2019privacy}. 
To date, a comparative analysis that simultaneously evaluate the utility of the three main families of DGMs does not exist. 
Chen et al.~\cite{chen2024synthetic} conducted a comparative analysis to evaluate the utility of GANs and DMs in enhancing weed recognition within the precision agriculture domain. 
Their study focused exclusively on these two families of DGMs and demonstrated the superior performance of DMs in this specific application.
Another significant trade-off involves fidelity and privacy, as typically, an increase in fidelity leads to a decrease in the privacy of synthetic data. Several approaches exist to assess the privacy of synthetic data~\cite{el2022validating, kossen2022toward,coyner2022synthetic}, yet determining the optimal strategy remains a challenging area of research. 
Some recent studies have compared the privacy-preserving ability of DMs and GANs, highlighting the greater tendency of DMs to memorize training set data~\cite{carlini2023extracting, akbar2023beware}.
Another important aspect warranting further investigation is the potential of synthetic data to enhance robustness against adversarial attacks. Recent efforts have aimed to benchmark the adversarial robustness of models trained on synthetic data. 
Sehwag et al.~\cite{sehwag2021robust} conducted a comparative analysis focusing exclusively on synthetic images generated by GANs and DMs, concluding that the latter offer superior robustness against adversarial perturbations. 

In contexts characterized by data scarcity, it becomes imperative to examine factors utility, privacy, and robustness, as these significantly affect the performance and reliability of DL models under such conditions.
The importance of a focused analysis in this context cannot be underestimated, as data scarcity often exacerbates challenges such as reduced model representational capacity and increased privacy risks, making it crucial to understand how these dimensions interact and affect overall outcomes. 
Ensuring utility, for instance, becomes more difficult due to the reduced representational capacity of small datasets, which can compromise the generalization capabilities of models~\cite{cao2022survey}. 
Similarly, maintaining robustness is more challenging when data are scarce, as models may be more susceptible to overfitting or adversarial vulnerabilities~\cite{xiong2024all}. Privacy concerns are also magnified, as smaller datasets increase the risk of exposing sensitive information~\cite{giomi10unified}. 
To mitigate these issues, approaches such as transfer learning ~\cite{weiss2016survey} and DA~\cite{chlap2021review} are widely used to improve utility by leveraging pre-trained models or generating synthetic data to augment the limited dataset, respectively.
Regarding robustness, methods like adversarial training~\cite{bai2021recent} have been employed to enhance model resilience by intentionally introducing perturbations during training to improve defenses against adversarial attacks.
For privacy, federated learning~\cite{zhang2021survey} allows distributed model training without centralizing sensitive data, thus reducing privacy risks. 
However, despite these advancements, the current state-of-the-art lacks a comprehensive evaluation that systematically addresses utility, robustness, and privacy in the context of data scarcity. 
This gap underscores the need for further research that simultaneously assesses these three critical dimensions under such constraints.

\section{Materials}
To evaluate the synthetic images generated by various DGMs, we utilized four public datasets: two from the medical domain and two from precision agriculture. 
This selection was motivated by the necessity to conduct analyses in domains where data scarcity poses a significant challenge, which synthetic data can potentially alleviate. In the medical domain, the scarcity of data may arise from various factors such as a limited number of patients for certain conditions or the absence of standardized protocols~\cite{leotsakos2014standardization}. Moreover, medical practices are constrained by privacy regulations~\cite{GDPR2016}, where compliance, in the absence of explicit consent, necessitates full anonymization of data to ensure adherence. Similarly, in precision agriculture, data scarcity stems from a combination of factors, including limited adoption of advanced technologies, lack of sensors, and various logistical challenges associated with data collection~\cite{soussi2024smart}.

The first medical dataset, Kvasir~\cite{bib:pogorelov2017kvasir}, comprises 8,000 endoscopic images divided into 8 classes that include anatomical references, pathological findings, and endoscopic procedures related to the gastrointestinal tract. Specifically, anatomical references include the Z-line, cecum, and pylorus, while pathological findings include esophagitis, polyps, and ulcerative colitis. Additionally, the Kvasir dataset provides two categories related to polyp removal: ``dyed and lifted polyp'' and ``dyed resection margins''.
The images have a resolution of $256\times256$.

The second medical dataset, CheXpert~\cite{irvin2019chexpert}, is a dataset of chest radiographs categorized into 14 different findings. For each pathology, one of three distinct labels was assigned: ``positive" (indicating the presence of the pathology in the image), ``negative" (indicating the absence of the pathology), or ``uncertain" (denoting insufficient information to ascertain the presence or absence of the pathology).
For the management of ``uncertain" labels, we adopted the U-Ones approach~\cite{irvin2019chexpert}, whereby all uncertain labels were treated as positive. For the generation task, we filtered the CheXpert dataset to include only antero-posterior images with a single pathology, excluding combinations of diseases to enable single class-conditional generation. After this filtering operation, a total of 40,447 chest radiographs were obtained.
Images are characterized by a resolution of $390\times320$. Despite the seemingly large number of images, when distributed across 14 classes, this dataset presents a scenario of data scarcity, with each class containing a limited number of samples, which is a common challenge in medical imaging tasks.

In the domain of precision agriculture, we employed the PlantVillage dataset~\cite{hughes2015open}, which contains 54,309 leaf images spanning 14 species which are divided into 38 classes.
The images have a resolution of $256\times256$.
Again, although the total number of images is considerable, the distribution across 38 classes results in a relatively low number of images per class, creating a data-scarce environment that mirrors real-world challenges in agricultural image analysis.
Finally, as a last dataset we used a subset of PlantVillage, focusing on Corn species. 
The Corn dataset comprises 3,852 images categorized into 4 distinct classes.
The images in this subset also have a resolution of $256\times256$, as it is derived from the PlantVillage dataset.
This subset was selected to explore the efficacy of DGMs across datasets characterized by varying numbers of classes, ranging from extensive to more limited class distributions.

Each dataset was partitioned into training, validation, and test sets using an 80\%-10\%-10\% hold-out method. We ensured that the class distribution was consistent across all three subsets. As preprocessing, each image was normalized to obtain pixel in the range [0,1]. 
In addition, the images from the CheXpert dataset were resized to a resolution of 256$\times$256 pixels.

\begin{table*}
\caption{Dataset characteristics}
\label{tab:TAB1}
\begin{adjustbox}{width=\textwidth}
\begin{tabular*}{\textwidth}{@{\extracolsep{\fill}} l l l l@{}}
\toprule
\textbf{Dataset} & \textbf{Domain} & \textbf{\# Images} & \textbf{\# Classes} \\
\midrule
Kvasir~\cite{bib:pogorelov2017kvasir} & Medical & 8 000  & 8 \\
CheXpert~\cite{irvin2019chexpert} & Medical & 40 447  & 14 \\
PlantVillage~\cite{hughes2015open}& Precision Agriculture & 54 309 & 38 \\
Corn~\cite{hughes2015open} & Precision Agriculture & 3 852 & 4 \\
\bottomrule
\end{tabular*}
\end{adjustbox}
\end{table*}

\section{Methods}
This section introduces the experimental framework developed to evaluate DGMs under conditions of data scarcity, extending the Generative Learning Trilemma. 
We begin by outlining six evaluation metrics: fidelity, diversity, sampling speed, utility, robustness, and privacy, which are detailed in Sections~\ref{subsubsec:Fidelity}-\ref{subsubsec:Privacy}. 
The specific experimental approaches associated with each metric are also discussed. 
Following this, Section~\ref{subsec:DGMs} presents the experimental setup, focusing on the three core generative backbones that form the foundation of our analysis.

\subsection{Metrics}

To introduce the metrics, hereinafter we adopt the following notation:
\begin{itemize}
    \item $G$ denotes a Deep Generative Model;
    \item $ \mathcal{X}$ represents the data space;
    \item $D_r=\{ r_1, \dots, r_n \} \subset \mathcal{X}$ is the real training dataset consisting of $n$ samples drawn from the real data distribution;
    \item $D_s = \{ s_1, \dots, s_m \} \subset \mathcal{X}$ is the synthetic dataset consisting of $m$ samples generated by $G$, where each synthetic sample $s_j$ is produced by $s_j = G(z_j)$ with $z_j$ sampled from a prior distribution.
\end{itemize}

\begin{figure*}[h]
	\centering
	\includegraphics[width=13.0cm]{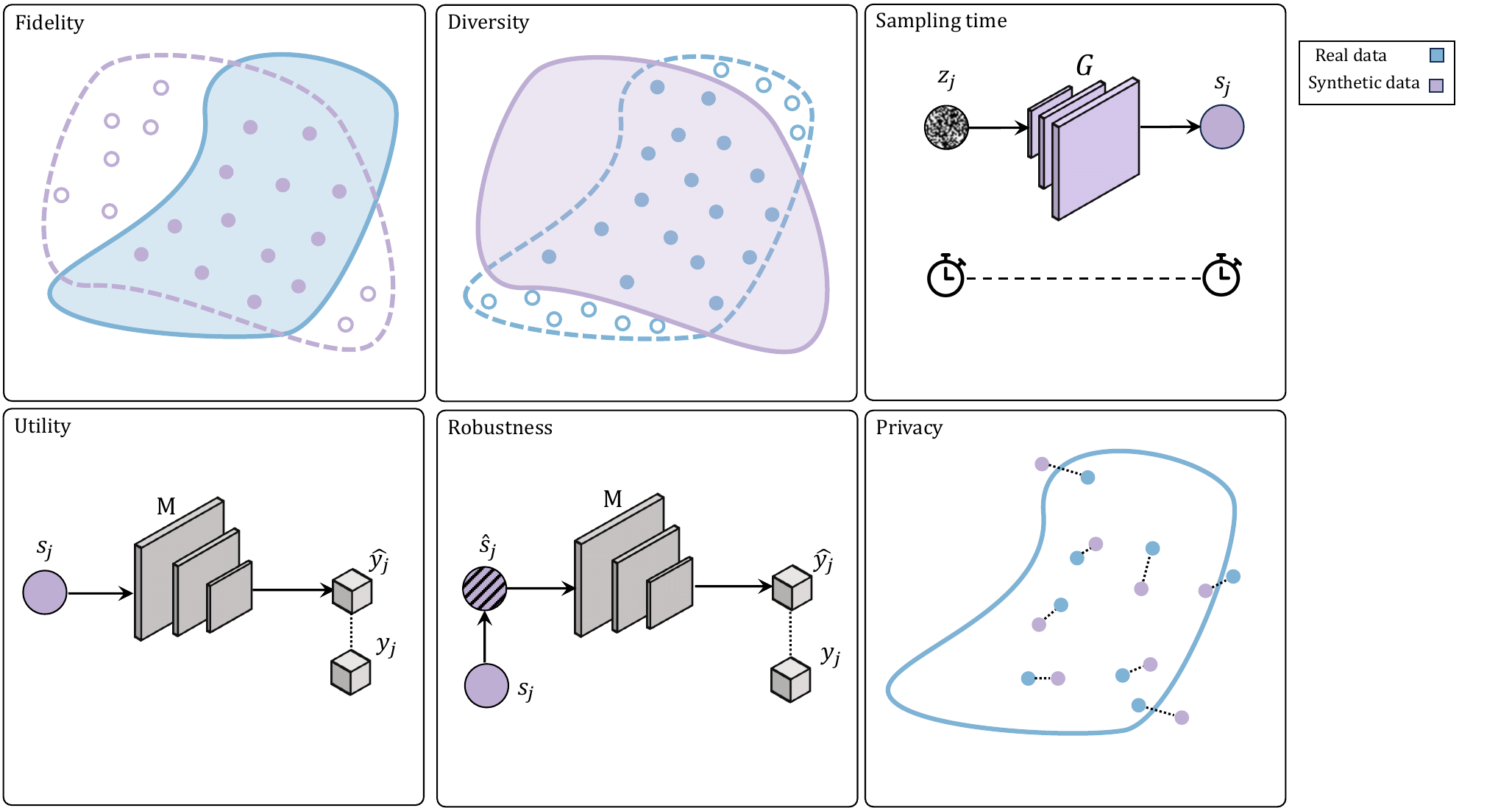}
	\caption{Conceptual overview illustrating the intuition behind the six metrics defined for evaluating DGMs: fidelity, diversity, sampling speed, utility, robustness, and privacy.}
	\label{FIG:1}
\end{figure*}

\subsubsection{Fidelity} \label{subsubsec:Fidelity}

Fidelity refers to the degree to which the generated samples accurately reflect the true data distribution. 
As illustrated in Fig.~\ref{FIG:1}, it assesses whether the generated samples lie within the actual data distribution, thereby avoiding unrealistic or low-quality outputs.
From a more formal perspective, we define that the fidelity $\mathcal{F}$ measuring the distance between the real data distribution and the synthetic data distribution.
To asses fidelity, we employ Precision~\cite{kynkaanniemi2019improved}, which quantifies the fraction of generated images that are realistic, focusing on how many of the generated samples lie within the manifold of real images. This metric provides a more direct measure of similarity between the real and synthetic data distributions. 
Higher Precision values indicate better fidelity, as more generated samples are close to real ones.
In this case, the Precision metric is calculated using a pretrained VGG-16 network~\cite{simonyan2014very}. 
Feature vectors $ \phi_r $ are extracted for each real image, and $ \phi_s $ for each synthetic image, resulting in two sets of feature vectors: $ \Phi_r $ and $ \Phi_s $. For each set, the corresponding manifolds are estimated in feature space by calculating the Euclidean distance between vectors. 
For each feature vector, a multidimensional hypersphere is formed with a radius equal to the distance from its $ k $-nearest neighbor (with $ k = 3 $ in this study). The Precision is then defined as:

$$
\mathcal{F}(G) = \frac{1}{|\Phi_s|} \sum_{\phi_s \in \Phi_s} f(\phi_s, \Phi_r)
$$
where the function $ f(\phi, \Phi) $ returns 1 if the feature vector $ \phi$ falls inside the hypersphere corresponding to the real data manifold, indicating that it is a valid representation of the data; otherwise, it returns 0, indicating that the vector does not belong to the manifold.
This ensures that the Precision metric captures whether each generated image is within the real data manifold.

\subsubsection{Diversity}\label{subsubsec:Diversity}
The diversity measures how well the synthetic data captures the variability of the real data distribution. 
As depicted in Fig.~\ref{FIG:1}, this metric evaluates whether the model can generate samples that span the entire range of possibilities within the true data, rather than being confined to a limited subset.
To evaluate the diversity, we employ the Recall~\cite{kynkaanniemi2019improved}, which measures the fraction of the real data distribution covered by the DGM. 
Similar to Precision, Recall is also calculated using the VGG-16 extracted feature vectors, and it is defined as:

$$
\mathcal{D}(G) = \frac{1}{|\Phi_r|} \sum_{\phi_r \in \Phi_r} f(\phi_r, \Phi_s)
$$
it checks if each real image is contained within the manifold of generated images, thereby measuring how much of the real data’s diversity is represented in the generated data.

In addition to using Precision to assess fidelity and Recall to measure diversity, we employed the Fréchet Inception Distance (FID)\cite{NIPS2017_8a1d6947} as an evaluation metric for synthetic image generation. 
While FID is widely adopted for assessing the fidelity of generated images, it inherently captures aspects of both fidelity and diversity. 
Specifically, FID quantifies the alignment between the synthetic and real data distributions, reflecting fidelity, while also accounting for the variability among generated samples, thereby providing insight into diversity. 
The metric is computed by comparing the distributions of real and generated images in the feature space of a pretrained Inception network\cite{szegedy2015going}. Mathematically, FID is defined as:

$$
\text{FID} = \| \mu_r - \mu_s \|_2^2 + \mathrm{Tr}\left( \Sigma_r + \Sigma_s - 2 \left( \Sigma_r \Sigma_s \right)^{1/2} \right)
$$
where $\mu_r$ and $\mu_s$ are the mean feature vectors of the real and synthetic datasets respectively, $\Sigma_r$ and $\Sigma_s$ are the covariance matrices of the real and synthetic datasets respectively, $\mathrm{Tr}(\cdot)$ denotes the trace of a matrix and $\|\cdot\|_2$ is the Euclidean norm.

\subsubsection{Sampling Speed}\label{subsec:SamplingTime}
The sampling speed metric $\mathcal{S}$ measures the efficiency of a DGM in generating synthetic samples. 
As shown in Fig.~\ref{FIG:1}, it measures the time required for the model to generate a sample $s_j$ from a latent vector $z_j$, providing insights into the computational performance of the model.
We compute the sampling speed as the number of samples generated per second:
$$
\mathcal{S}(G) = \frac{c}{t}
$$
where $c$ is the number of samples generated, $t$ is the total time taken to generate these samples, therefore $\mathcal{S}(G)$ is measured in samples per second. 
In our scenario, a higher value of this metric indicates greater efficiency, making the model more suitable for real-world applications that require rapid generation of synthetic data.
\begin{figure*}[htb]
	\centering
	\includegraphics[width=13.0cm]{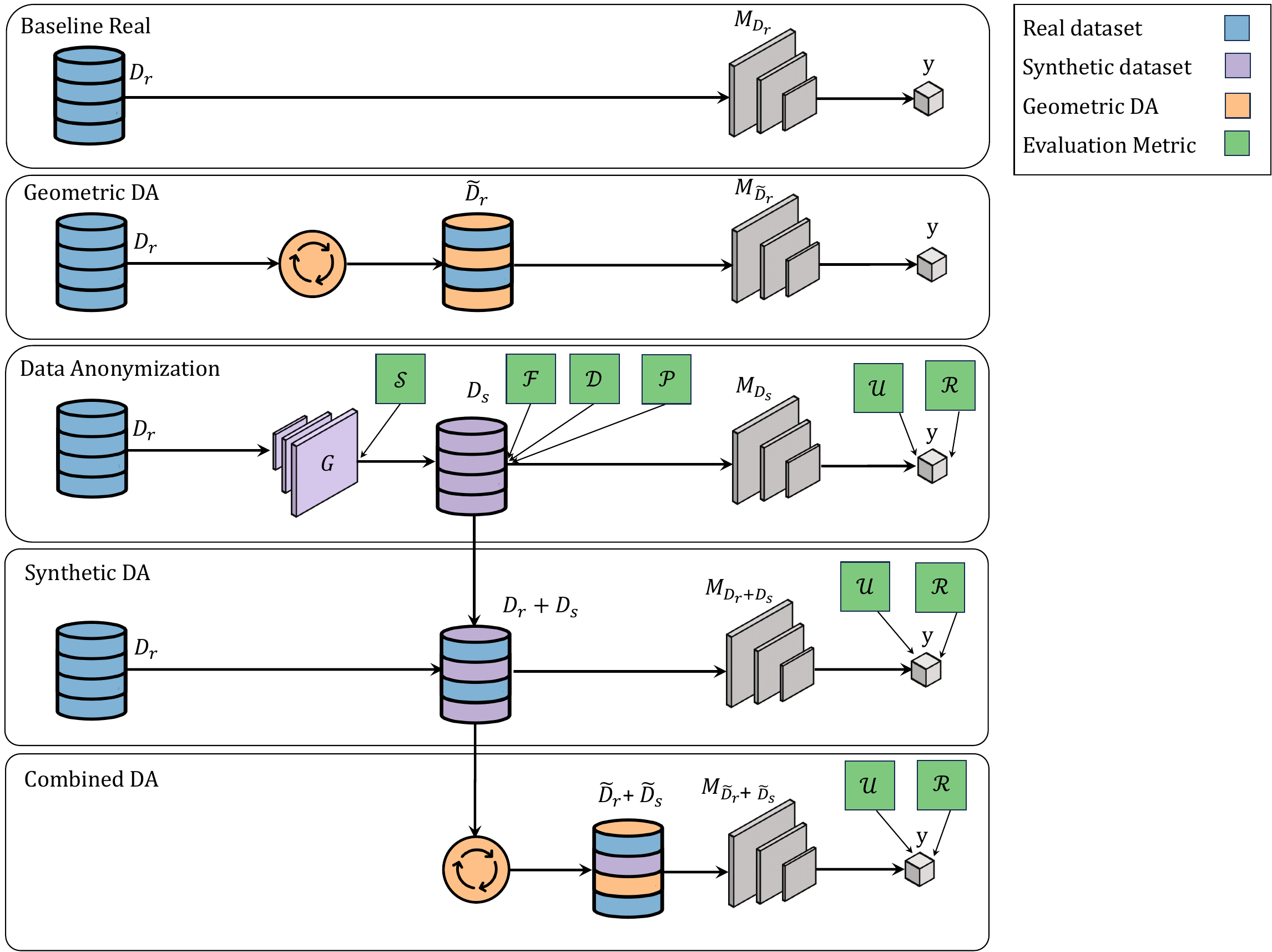}
	\caption{Overview of the methodology for the classification downstream task used to assess utility and robustness. Two baseline approaches are considered: one using real data and another incorporating Geometric DA for classifier training. Data Anonymization is then performed using only synthetic data generated by three deep generative models (DGMs): Variational Autoencoder (VAE), Generative Adversarial Network (GAN), and Diffusion Model (DM). In this phase, sampling speed (\(\mathcal{S}\)), fidelity (\(\mathcal{F}\)), diversity (\(\mathcal{D}\)), and privacy (\(\mathcal{P}\)) are also evaluated. Finally, Synthetic DA combines real data with synthetic samples from the DGMs, while Combined DA further integrates Geometric DA.}

	\label{FIG:2}
\end{figure*}

\subsubsection{Utility}\label{subsec:Utility}
The rationale behind the development of the utility metric is to assess the potential benefits of using synthetic data in real-world applications. 
For instance, as illustrated in the Fig.~\ref{FIG:1}, utility can be evaluated through the performance of a supervised model $M_{D_s}$ trained on synthetic samples $D_s$ and tested on a real test set $D^T_r$.
The utility is defined as:
$$
\mathcal{U}(G, M_{D_s}, D^T_r) = A(M_{D_s}, D^T_r)
$$
where $A(\cdot, \cdot)$ is a performance metric (i.e., accuracy, precision, recall, F1-score) used to assess $M_{D_s}$ on $D^T_r$.

To evaluate the utility of synthetic images generated by DGMs, we designed a series of classification tasks, structured to explore the effect of DA techniques that leverage synthetic images. 
In these experiments, we use classification accuracy as performance metric $A(\cdot, \cdot)$ to compare different approaches, systematically varying the training sets to assess how synthetic data impacts model performance.
Such experiments were repeated for each DGM considered for analysis and for each of the 4 datasets, using some DL architecture and hyperparameters, detailed in~\ref{appendixA}, to ensure consistent and reproducible results. 
They are:

\begin{enumerate}
    \item \textbf{Baseline Real:} in the first classification task, shown in the uppermost panel of Fig.~\ref{FIG:2}, $M_{D_r}$ is trained only on the real dataset $D_r$. 
    Its performance is evaluated on the real test set $T_r$ to establish a baseline for later investigating the impact of synthetic data.

    \item \textbf{Geometric Data Augmentation:} this experiments investigates the impact of traditional DA techniques, by applying a set of classical DA techniques that include 3 affine transformations (rotations, horizontal and vertical flips), which produces an augmented set $\tilde{D}_r$. This set, as shown in the second panel of Fig.~\ref{FIG:2}, is then used to evaluate the potential performance improvements given by these simple transformations.

    \item \textbf{Data Anonymization:} this experiment aims to evaluate the performance of models trained solely on synthetic data, a useful approach in scenarios where data anonymization is required to protect sensitive information. The experiments were designed to assess how effectively synthetic datasets could substitute real training data.
    To achieve this, we conducted three experiments using fully synthetic datasets $D_s$. 
    As illustrate in Fig.~\ref{FIG:2}, these datasets were generated by a class-conditional DGM, trained to produce images corresponding to specific input labels. The synthetic datasets were created in three different sizes: (i) one matching the number of images in the real training set, (ii) another with twice as many images, and (iii) a third with three times as many images. 
    For clarity, Fig.~\ref{FIG:2} presents only the experiment conducted with a synthetic dataset of the same cardinality as the actual training set.
    
    For each case, the utility of the model trained on $D_s$ was evaluated as:
    
    $$
    \mathcal{U}(G, M_{D_s}, D^T_r) = A(M_{D_s}, D^T_r)
    $$

    \item \textbf{Synthetic Data Augmentation:} this set of experiments aims to assess potential performance improvements by adding synthetic data to real data. 
    As shown in the Synthetic DA panel of Fig.~\ref{FIG:2}, the study investigates the impact of augmenting real datasets with varying amounts of synthetic data, creating augmented datasets of different sizes. 
    Three augmented datasets were formed: (i) one by adding a number of synthetic images equal to the size of the real training set, (ii) one by adding double that size, and (iii) another by adding triple that size of synthetic images. 
    The models were trained on these combined datasets, and their utility was calculated:

    $$
    \mathcal{U}(G,M_{D_r+D_s}, D^T_r) = A(M_{D_r+D_s}, D^T_r)
    $$

    \item \textbf{Combined Data Augmentation:} in the last set of experiments illustrated in Fig.~\ref{FIG:2}, we analyzed the impact of applying Geometric DA techniques on datasets comprising both real and synthetic data. 
    Specifically, geometric transformations were utilized to generate augmented versions of the data, resulting in the creation of three distinct datasets: (i) a dataset combining the augmented real training set with an augmented synthetic dataset of equivalent size, (ii) a dataset consisting of the augmented real training set and an augmented synthetic dataset twice the size of the real training set, and (iii) a dataset that combines the augmented real training set with an augmented synthetic dataset three times the size of the real training set.
    The utility in this case was computed as:
    $$
    \mathcal{U}(G,M_{\tilde{D}_r+\tilde{D}_s}, D^T_r) = A(M_{\tilde{D}_r+\tilde{D}_s}, D^T_r)
    $$

\end{enumerate}

\subsubsection{Robustness}\label{subsubsec:Robustness}
The introduction of this score is motivated by the need to evaluate the robustness of models trained on synthetic data when subjected to adversarial perturbations. 
As illustrated in Fig.~\ref{FIG:1}, the robustness metric $R$ evaluates the resilience of a model $M_{D_s}$, trained on synthetic data $D_s$, against an adversarial attack using a perturbed version $\hat{s}_j$ of the sample:

$$
\mathcal{R}(G, M, \delta) = A(M_{D_s}, \delta(D^T_r))
$$
where $\delta(D^T_r)$ is the real test set $D^T_r$ after applying a perturbation $\delta$ and $A(M_{D_s}, \delta(D^T_r))$ measures the performance of the perturbed model on the real test set $D^T_r$.
For each of the experiments shown in the Fig.~\ref{FIG:2}, in a manner equal to the utility, we calculated the robustness by varying the training sets and measuring the impact of the adversarial perturbation on the model accuracy.
To simulate it, we applied adversarial attacks using the DeepFool algorithm~\cite{moosavi2016deepfool} to the inputs of $M_{D_s}$, and we measured the adversarial accuracy of the perturbed model on the real test set $D^T_r$. 
DeepFool is an efficient technique designed to compute the minimal perturbations required to mislead neural networks. 
It operates by iteratively calculating the smallest perturbation needed to move an input sample across the decision boundary of a classifier, causing a misclassification. 
To maintain uniformity in our tests across different models, we consistently implemented the DeepFool attack with a fixed number of five iterations.

\subsubsection{Privacy}\label{subsubsec:Privacy}
The need for a metric to assess the privacy of DGMs is driven by the importance of evaluating the similarity between synthetic and real datasets. 
The synthetic distribution must closely approximate the real distribution while maintaining sufficient distance between individual synthetic and real data points to safeguard sensitive information. 
In fact, a greater separation between synthetic and real samples corresponds to enhanced privacy safeguards. 
However, there is a need to balance privacy and fidelity: for high fidelity, synthetic samples should closely approximate real samples, while privacy requires some distance to protect sensitive data.
As is shown in Fig.~\ref{FIG:2}, this approach is based on the calculation of distances between real and synthetic data points.
The privacy metric $\mathcal{P}$ is designed to evaluate the risk of sensitive information from the real dataset $D_r$ being leaked through the synthetic dataset $D_s$. 
In our analysis, we addressed the concern of privacy risks by implementing the methodology presented by Akbar et al.~\cite{akbar2023beware}. 
Specifically, for each synthetic set, we randomly selected $q$ synthetic images. 
Each synthetic image was then compared to all images in the corresponding real training set by calculating the Structural Similarity Index (SSIM). 
For each synthetic image $s_j$, we recorded the highest correlation with any real image $r_i$. This resulted in the formulation:

$$
\mathcal{P}_j = \max_{r_i \in D_r} \text{SSIM}(s_j, r_i)
$$
The Privacy metric, $\mathcal{P}(G)$, is subsequently computed as the average of these maximum similarities $\mathcal{P}_j$ across the synthetic images. Due to computational demands, we limited the analysis to 100 synthetic images. However, to enhance the robustness of the analysis, we repeated this extraction 10 times.

$$
\mathcal{P}(G) = \frac{1}{l} \sum_{i=1}^{l} \left( \frac{1}{q} \sum_{j=1}^{q} \mathcal{P}_{j} \right)
$$
where $n$ is the total number of synthetic images selected for the analysis and $l$ is the total number of extraction attempts. Being a measure of similarity, a lower value of this metric indicates a higher level of privacy.
This metric is particularly crucial in sensitive applications such as the medical field, where identifying which DGMs are more prone to generating synthetic images that closely resemble real patient data is essential to prevent potential privacy breaches.

\subsection{Deep Generative Models} \label{subsec:DGMs}
In this section, we provide a detailed description of the experimental setup used to implement the class-conditional DGMs chosen for our analysis: VAE, GAN, and DM. 
For the family of GANs, we used StyleGAN2 (SG2), while for the DMs, we used the Latent Diffusion Model (LDM). 
Each model was selected to represent a state-of-the-art approach within their respective families: VAE was chosen for its strong probabilistic foundation and ability to model complex data distributions, SG2 for its superior image synthesis capabilities and architectural innovations within the GAN family, and LDM for its efficiency and scalability in generating high-quality samples within the DM framework. 
Although DGMs have demonstrated impressive performance, it should be pointed out that they generally require large amounts of data to achieve optimal results. 
For this reason, in many applications like medicine and precision agriculture, data scarcity is a significant challenge.
Evaluating the performance of these models with limited data is crucial for identifying strategies to maintain their effectiveness. 
So, the aim of this assessment is to identify which DGMs are most suitable for low-data environments.

\subsubsection{Variational Autoencoder}
\begin{figure*}
	\centering
	\includegraphics[width=10.0cm]{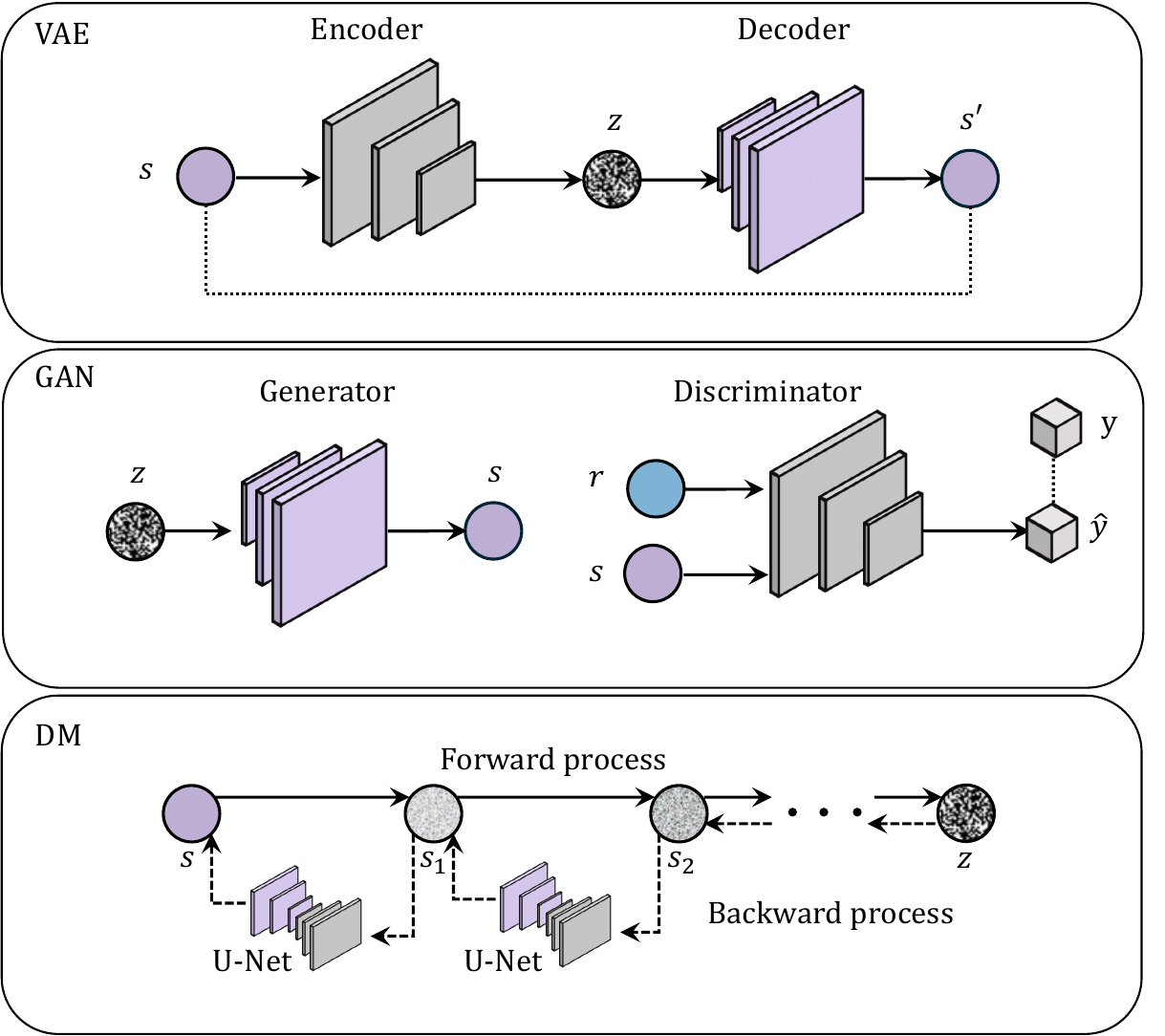}
	\caption{Architectures of Deep Generative Models. (Top) VAE comprising an encoder that maps input data $s$ to a latent variable $z$, and a decoder reconstructing the output $s'$. (Middle) GAN including a generator, which maps latent variable $z$ to synthetic data $s$, and a discriminator distinguishing real data $r$ from generated samples. (Bottom) DM illustrating the forward process that adds noise to data $s$ iteratively to generate $s_1, s_2, \dots$, and a backward process that learns to reverse this chain, obtaining latent representation $z$.}
	\label{FIG:3}
\end{figure*}

VAEs~\cite{kingma2013auto} are a type of DGM that combine neural networks with variational Bayesian methods to model complex data distributions and generate new data. Unlike traditional AEs, VAEs use a probabilistic latent space, allowing them to learn a distribution over latent variables, which enhances their generative abilities. As illustrated in Fig.~\ref{FIG:3}, the VAE architecture comprises an encoder, which maps the input sample $s$ into a latent vector $z$, and a decoder, which generates a reconstruction of the input sample, denoted as $s'$, from the latent representation $z$.
We implemented a Conditional Variational Autoencoder (CVAE)~\cite{sohn2015learning}, where the encoder architecture was composed of a sequence of convolutional layers with hidden dimensions of 16, 32, 54, 128, 256, and 1024. Each convolutional layer was followed by batch normalization and a LeakyReLU activation function~\cite{xu2020reluplex}. The encoder's output was passed through two linear layers to estimate the mean and log-variance of the latent space, with the latent dimension fixed at 128. The decoder network mirrored the encoder's structure, and the condition vector was concatenated with the input data for the encoder and with the sampled latent vector for the decoder.
The proposed architecture results in a model with an order of magnitude of tens of millions of parameters.
We trained the CVAE by minimizing the standard loss which consist of a reconstruction loss and a Kullback-Leibler divergence term, balanced by a $\beta$ factor of 0.002 in our scenario. Training utilized the Adam optimizer, with a learning rate of 0.001 and a weight decay of 0.01. The model was trained for 100 epochs.

\subsubsection{StyleGAN 2}
GANs are a class of DGMs designed to generate realistic images through
an adversarial process, where a generator creates synthetic images, and a discriminator differentiates between real and fake images. 
Fig.~\ref{FIG:3} depicts the architecture of a GAN. The generator takes a latent noise vector $z$ as input and transforms it into a synthetic data sample $s$, which is designed to resemble real data.
The discriminator, on the other hand, is trained to distinguish between real images $r$ and synthetic images $s$ generated by the generator. The discriminator outputs a prediction $\hat{y}$, which is compared to the true label $y$ indicating whether the input is real or fake.
Through this adversarial process, the generator learns to improve the quality of the synthetic images, while the discriminator becomes more able to distinguishing real images from the generated ones.

SG2~\cite{karras2020analyzing} builds on this foundation with several innovations to enhance image quality. One major improvement is the introduction of a Mapping Network, which transforms the latent code into an intermediate space, offering greater control over image attributes. SG2 also employs Adaptive Instance Normalization (AdaIN), modulating the generator's layers to adjust image styles. Additional techniques, like Weight Demodulation, Style Mixing Regularization, and Path Length Regularization, help improve image coherence, detail, and overall fidelity, making SG2 a leading model for high-quality image generation.
We implemented a conditional version of the SG2 architecture as described by Karras et al.~\cite{karras2020analyzing} to generate high-resolution images based on class labels. The generator network employed AdaIN and a Mapping Network to transform latent codes and class embeddings into style vectors. The Mapping Network comprised 8 layers, with both the input latent dimension and the intermediate latent space dimension set to 512.
The total number of parameters in our implementation is on the order of 50 millions.
A fixed batch size of 32 was utilized throughout the training process. The Adam optimizer was applied to both the generator and discriminator, with a learning rate of 0.0025, betas of [0, 0.99], and epsilon set to $1 \times 10^{-8}$. The training loss adhered to the non-saturating logistic formulation, incorporating R1 regularization and path length regularization. The model was trained for 100 epochs.

\subsubsection{Latent Diffusion Model}
DMs~\cite{denoisingDiffusionProbModels} are a new class of DGMs known for generating highly realistic synthetic data. The operation of these models relies on two key processes: the forward process and the reverse process. As described in Fig.~\ref{FIG:3}, the forward process starts from a real sample $s$, progressively adding noise to generate increasingly noisy versions $ s_1, s_2, \dots $ until reaching complete noise represented by $ z $. The reverse process, learned by a neural network, performs the opposite task, gradually removing noise from $ z $ to recover the original sample $ s $.
Despite their effectiveness, DMs require significant training and sampling time. To address this, LDMs~\cite{high_resolutinLDM} were introduced. LDMs operate in a compressed latent space instead of pixel space, significantly reducing computational costs while maintaining performance, as the model learns to denoise the latent representation of the data.
Our implementation of LDM follows the approach outlined by Rombach et al.~\cite{high_resolutinLDM}. Initially, an AE~\cite{esser2021taming} was employed to learn the latent space in which the DM was subsequently trained. The architecture of both the encoder and decoder mirrors each other, consisting of a sequence of ResNet blocks~\cite{he2016deep}. The implementation of the AE loss strictly followed the approach outlined by Rombach et al.~\cite{high_resolutinLDM}. A $\beta$ factor of $10^{-6}$ was utilized to ensure high-fidelity reconstruction. The latent space was defined with a dimensionality of $64 \times 64 \times 3$, while a learning rate of $4.5 \times 10^{-6}$ and a batch size of 16 were implemented. Each AE for the respective datasets was trained for 50 epochs.
Subsequently, the weights of the AE were employed to train the DM. The model accepts input images along with their corresponding labels. The input images are compressed into latent vectors by the AE, while the labels are encoded into embedding vectors using a lookup table, with the dictionary size corresponding to the number of classes. The resulting embedding vectors have a dimensionality of 512 and serve to provide conditional information for class generation through the cross-attention mechanism.
Denoising U-Net~\cite{ronneberger2015u} is characterized by a succession of modules consisting of a ResNet block and an attention layer. For training the LDM, the standard loss, specified in~\cite{high_resolutinLDM}, was implemented. 
The total parameter count in our LDM implementation ranges between 400 million to 600 million, depending on the dataset and class embeddings.
The following parameters were utilized: a learning rate of $10^{-6}$, a batch size of 16, and a linear noise schedule. The initial noise parameter $\beta_0$ was set to 0.0015, while the final noise parameter $\beta_T$ was set to 0.0195. The training process was conducted over 1000 timesteps.
The LDMs for each dataset were trained for 100 epochs.

\section{Results and Discussions}

This section presents a comprehensive evaluation of the DGMs across the six metrics. The analysis begins with an assessment of the Generative Learning Trilemma metrics, providing insights into the trade-offs between fidelity, diversity, and sampling speed, alongside an evaluation of privacy preservation. 
This is followed by an investigation of downstream classification performance, examining the utility and robustness of synthetic data when used for training. 
The impact of increasing the quantity of synthetic data is then analyzed to determine its influence on utility and robustness. 
Finally, a comparative analysis summarizes the strengths and limitations of each model, offering a broader perspective on their applicability across different datasets and domains.

\subsection{Evaluation of Generative Learning Trilemma Metrics and Privacy}

\begin{table*}[htbp]
\caption{Results of metrics assessing fidelity, diversity, privacy, and sampling speed.}
\label{tab:TAB2}
\begin{adjustbox}{width=\textwidth} 
\begin{tabular}{l l l l l l l l}
\toprule
\textbf{Dataset} & \textbf{Models} & \textbf{Fidelity $\uparrow$} & \textbf{Diversity $\uparrow$} & \textbf{FID $\downarrow$} & \textbf{Privacy $\downarrow$} & \textbf{Sampling Speed $\uparrow$} \\
\midrule
 & VAE & \num{1.50e-02} & \num{0.00e+00} & 266.32 & \num{7.20e-01} & \textbf{12.80} \\
Kvasir & GAN & \num{2.67e-01} & \num{2.00e-04} & 159.76 & $\mathbf{5.27 \times 10^{-1}}$ & 8.00 \\
 & DM & $\mathbf{4.49 \times 10^{-1}}$ & $\mathbf{1.44 \times 10^{-1}}$ & \textbf{37.61} & \num{5.39e-01} & 0.16 \\
\midrule
 & VAE & \num{6.30e-02} & \num{0.00e+00} & 383.92 & \num{6.49e-01} & \textbf{12.80} \\
CheXpert & GAN & \num{3.01e-01} & \num{4.00e-04} & 50.87 & $\mathbf{4.13 \times 10^{-1}}$ & 8.00 \\
 & DM & $\mathbf{3.89 \times 10^{-1}}$ & $\mathbf{8.95 \times 10^{-2}}$ & \textbf{20.65} & \num{4.59e-01} & 0.16 \\
\midrule
 & VAE & \num{7.40e-03} & \num{0.00e+00} & 338.64 & \num{7.13e-01} & \textbf{12.80} \\
Corn & GAN & \num{3.41e-01} & \num{3.00e-04} & 83.37 & $\mathbf{3.68 \times 10^{-1}}$ & 8.00 \\
 & DM & $\mathbf{5.71 \times 10^{-1}}$ & $\mathbf{1.06 \times 10^{-1}}$ & \textbf{38.28} & \num{3.93e-01} & 0.16 \\
\midrule
 & VAE & \num{2.00e-04} & \num{0.00e+00} & 286.31 & \num{7.27e-01} & \textbf{12.80} \\
PlantVillage & GAN & $\mathbf{5.62 \times 10^{-1}}$ & $\mathbf{3.51 \times 10^{-1}}$ & \textbf{26.88} & $\mathbf{3.27 \times 10^{-1}}$ & 7.10 \\
 & DM & \num{4.76e-01} & \num{1.57e-01} & 39.9 & \num{3.79e-01} & 0.16 \\
\hline
\hline % Double line before average rows
 & VAE & \num{2.14e-02} & 0.00 & 318.80 & \num{7.02e-01} & \textbf{12.80} \\ 
Average & GAN & \num{3.68e-01} & \num{8.79e-02} & 80.22 & $\mathbf{4.09 \times 10^{-1}}$ & 8.00 \\ 
 & DM & $\mathbf{4.72 \times 10^{-1}}$ & $\mathbf{1.12 \times 10^{-1}}$ & \textbf{34.11} & \num{4.42e-01} & 0.16 \\ 
\bottomrule
\end{tabular}
\end{adjustbox}
\end{table*}

Table \ref{tab:TAB2} presents the results involving the metrics that investigate the requirements introduced in the Generative Learning Trilemma~\cite{xiao2021tackling}, consisting of fidelity, diversity, and sampling speed, in the context of data scarcity. 
The table also includes the FID, which provides a combined measure of fidelity and diversity, as well as the results for privacy.
As can be seen from the average results, DM has the best fidelity, indicating highly realistic image generation, and the best diversity, suggesting a greater variety of generation. This is also confirmed by the lower average FID value.
In contrast, the low fidelity and null diversity, confirmed by the high FID reveal the limitations of VAE in generating large realistic images, as argued in the existing literature~\cite{bond2021deep}. 
In this case, VAE fails to train effectively and produces very similar and low quality images. 
Regarding the privacy, VAE exhibited the highest correlation with real images. 
This high correlation suggests that VAE is generating synthetic images that are very similar at the pixel level to the original images but lack realism. 
This observation supports the hypothesis, according to which VAE tends to produce repetitive and low-quality images rather than diverse and realistic ones.
GAN demonstrates a lower correlation compared to DM, suggesting better privacy preservation. 
The sampling speed revealed a significant limitation of DMs, as they require substantially more time to generate images compared to GAN and VAEs. 
Finally, in the comparison between GAN and VAE, the latter demonstrated an even higher sampling speed.

\begin{figure*}[t]
	\centering
	\includegraphics[width=1\columnwidth]{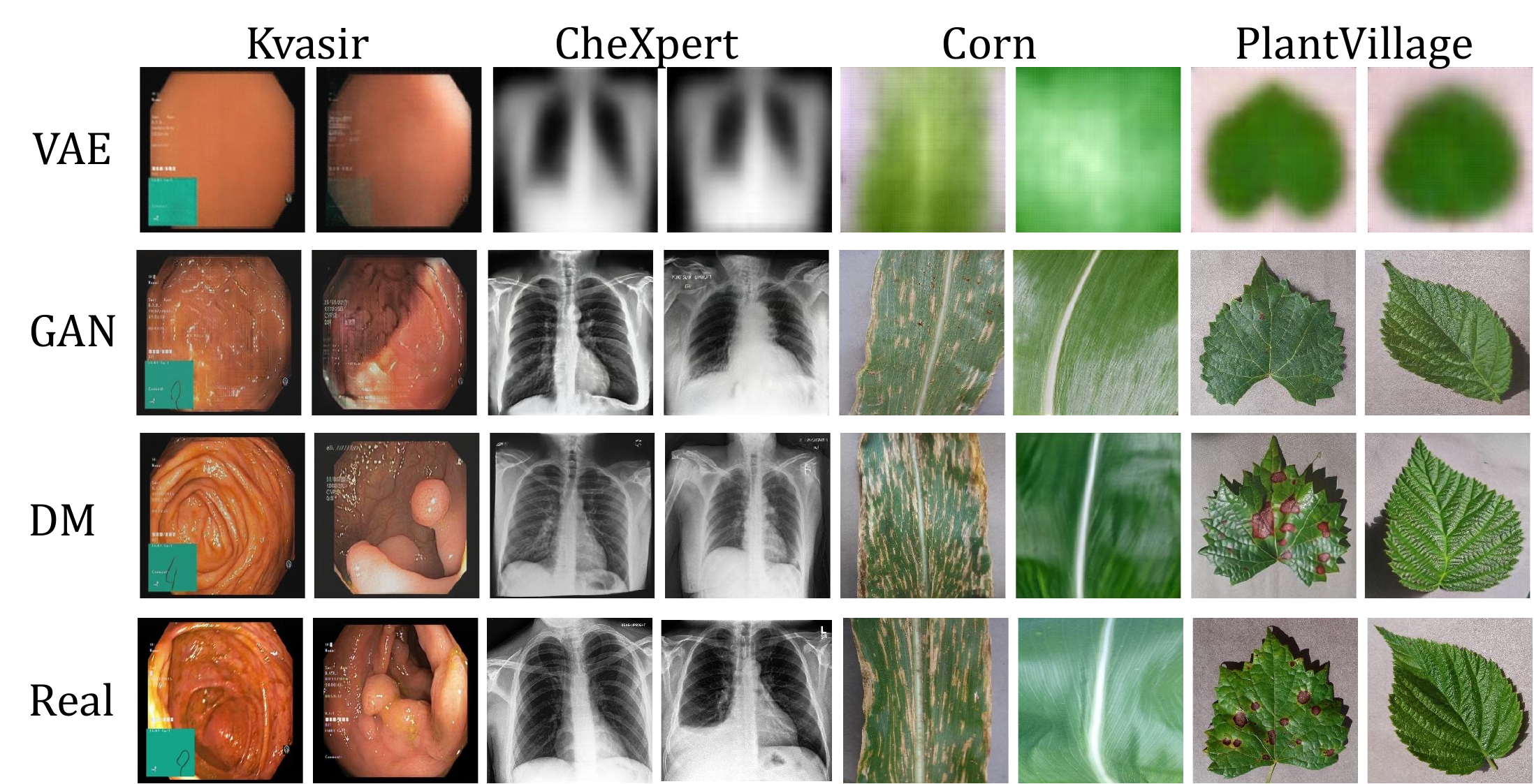}
	\caption{Comparison of synthetic images generated by DGMs with real images. For each of the four datasets used in the analysis, two images are presented.}
	\label{FIG:4}
\end{figure*}

Analyzing the four datasets individually reveals that general trends are present in the first three datasets, whereas the PlantVillage dataset warrants a separate analysis. 
For this dataset, GAN has demonstrated its ability to effectively scale and maintain a highly disentangled representation space, thereby accurately capturing the distinctive features of various classes within large datasets~\cite{wu2021stylespace}. 
Specifically, GAN achieves better fidelity and diversity compared to even DM.
For the other datasets, the improved fidelity and diversity observed in DMs compared to GAN and VAEs can be attributed to fundamental architectural and methodological differences. 
Unlike VAEs and GAN, DMs utilize a diffusion process that gradually denoises image representations over multiple steps. 
This progressive refinement enables DMs to capture fine-grained details and model complex data distributions.
Across all datasets, GAN demonstrates a lower correlation compared to DM, indicating better privacy preservation. 
This phenomenon can likely be attributed to the fact that, unlike the forward process in DMs, the GAN generator does not have direct access to the actual training images during its training phase.
Lastly, it should be noted that the sampling speed for DMs is inherently slower than that of VAEs and GANs, due to the iterative nature of the diffusion process, which involves multiple computational steps per sample.

Fig.~\ref{FIG:4} illustrates a comparison of the images generated by the three DGMs. Synthetic images of each of the four datasets are shown and compared with the corresponding real images. 
As shown, GAN and DM produce images of high visual quality, effectively capturing fine-grained details and achieving realistic textures. 
In contrast, the images generated by the VAE exhibit notable limitations, particularly in rendering high-resolution features, which results in a comparatively lower quality and a lack of sharpness. 
This highlights the differences in the capacity of these models to generate high-fidelity outputs.

\subsection{Downstream Classification Performance: Utility and Robustness}

\begin{table*}[h]
\caption{Results of metrics assessing utility and robustness for Kvasir and CheXpert.}
\label{tab:TAB3}
\centering
\begin{adjustbox}{width=\textwidth}
\begin{tabular}{@{}llcccccccc@{}}
\toprule
\multirow{2}{*}{\textbf{Experiments}} & \multirow{2}{*}{\textbf{Models}} & \multicolumn{2}{c}{\textbf{Kvasir}} & \multicolumn{2}{c}{\textbf{CheXpert}} \\
 &  & \textbf{Utility} & \textbf{Robustness} & \textbf{Utility} & \textbf{Robustness} \\
\midrule
Baseline Real & - & 0.804 & 0.243 & 0.687 & 0.684 \\
\midrule
Geometric DA & - & 0.814 & 0.225 & 0.672 & 0.669 \\
\midrule
\multirow{3}{*}{Data Anonymization} & VAE & 0.522 & 0.260 & 0.334 & 0.322 \\
 & GAN & 0.319 & 0.190 & 0.298 & 0.282 \\
 & DM & 0.559 & 0.253 & 0.519 & 0.515 \\
\midrule
\multirow{3}{*}{Synthetic DA} & VAE & 0.756 & 0.255 & 0.616 & 0.601 \\
 & GAN & 0.759 & \textbf{0.267} & 0.457 & 0.431 \\
 & DM & 0.860 & 0.195 & 0.689 & 0.688 \\
\midrule
\multirow{3}{*}{Combined DA} & VAE & 0.789 & 0.226 & 0.605 & 0.599 \\
 & GAN & 0.801 & 0.227 & 0.453 & 0.440 \\
 & DM & \textbf{0.875} & 0.154 & \textbf{0.720} & \textbf{0.718} \\
\bottomrule
\end{tabular}
\end{adjustbox}
\end{table*}

\begin{table*}[h]
\caption{Results of metrics assessing Utility and Robustness for Corn and PlantVillage.}
\label{tab:TAB4}
\centering
\begin{adjustbox}{width=\textwidth}
\begin{tabular}{@{}llcccccccc@{}}
\toprule
\multirow{2}{*}{\textbf{Experiments}} & \multirow{2}{*}{\textbf{Models}} & \multicolumn{2}{c}{\textbf{Corn}} & \multicolumn{2}{c}{\textbf{PlantVillage}} \\
 &  & \textbf{Utility} & \textbf{Robustness} & \textbf{Utility} & \textbf{Robustness} \\
\midrule
Baseline Real & - & 0.917 & 0.670 & 0.918 & 0.160 \\
\midrule
Geometric DA & - & 0.901 & 0.538 & \textbf{0.967} & 0.161 \\
\midrule
\multirow{3}{*}{Data Anonymization} & VAE & 0.536 & 0.525 & 0.145 & 0.105 \\
 & GAN & 0.746 & 0.644 & 0.619 & 0.208 \\
 & DM & 0.756 & 0.413 & 0.482 & 0.214 \\
\midrule
\multirow{3}{*}{Synthetic DA} & VAE & 0.912 & \textbf{0.767} & 0.866 & 0.178 \\
 & GAN & 0.909 & 0.765 & 0.877 & 0.205 \\
 & DM & 0.925 & 0.657 & 0.887 & 0.189 \\
\midrule
\multirow{3}{*}{Combined DA} & VAE & 0.894 & 0.536 & 0.920 & 0.172 \\
 & GAN & 0.880 & 0.705 & 0.952 & \textbf{0.238} \\
 & DM & \textbf{0.961} & 0.429 & 0.949 & 0.225 \\
\bottomrule
\end{tabular}
\end{adjustbox}
\end{table*}

Tab.~\ref{tab:TAB3} and Tab.~\ref{tab:TAB4} present the results obtained by DGMs in the downstream classification task, introduced to evaluate the utility and robustness of synthetic images. 
Overall, the applications of Geometric DA do not always derive benefits compared to the baseline with only real. 
This indicates that for some datasets suffering from overfitting, it is essential to have a larger as well as more diverse training set~\cite{ying2019overview}. 
As can be seen from the Data Anonymization experiments, the quality of the images generated still does not allow training with only synthetics to achieve performance comparable to that obtained using only real.
Simply adding synthetic images to the real training set did not produce significant improvements in performance.
While, the substantial improvement in classification performance was achieved by combining the addition of synthetic images with the application of Geometric DA. 
Regarding robustness, the results demonstrate a significant advantage conferred by the use of synthetic images. 
Models trained with synthetic images proved to be more robust to adversarial attacks. 
While no specific DGM was identified as generating synthetic images that provide superior robustness over others, all classifiers trained with synthetic images exhibited greater robustness compared to those trained exclusively with real images.
This underscores the potential of synthetic images to enhance the robustness of DL models against adversarial attacks, particularly in contexts where data scarcity is a significant challenge. 
Synthetic images can effectively augment limited datasets, providing a valuable resource for improving model performance and resilience.

Analyzing the Kvasir dataset specifically (Tab.\ref{tab:TAB3}), the application of Geometric DA alone led to only a 1\% increase in accuracy compared to the baseline. 
The highest performance among the Data Anonymization experiments was achieved by DM, which aligns with the best fidelity results.
Interestingly, for this dataset, the synthetic images generated by VAE resulted in higher utility than those produced by GAN, despite a significantly lower fidelity score. 
This suggests that, while VAE may not generate images with excellent visual quality, it effectively captures features that are useful for distinguishing between real classes.
This observation is crucial as it suggests that FID may not always be a reliable metric for assessing the utility of synthetic images in downstream tasks~\cite{jayasumana2024rethinking}. 
For this dataset, incorporating synthetic images generated by DM into the real training set yielded a significant improvement in performance, resulting in an accuracy increase of up to 6\% compared to the baseline using only real images.
The best overall performance was obtained by combining DM-generated synthetic images with Geometric DA, which led to a 7\% improvement in accuracy over the real baseline.

The classification of CheXpert, whose results are reported in Tab.\ref{tab:TAB3}, has proven to be more challenging compared to the others, given that even the experiment using only real images, employed as a reference baseline, exhibited very low performance. 
The reason for such low performance could be due to the exclusion of images of patients with multiple pathologies, which reduced the dataset's dimensionality. 
Another possible factor is the low quality of the available images, where pixels are represented as 8-bit integers. 
Given these constraints, we report the results for this dataset using top-3 accuracy and top-3 adversarial accuracy as metrics for evaluating utility and robustness.
In this dataset, the use of Geometric DA did not yield any significant improvement. 
Consistent with previous findings, synthetic images generated by the VAE outperformed those produced by GAN. 
The inclusion of synthetic images into the real training set did not lead to gains, except for a slight improvement in adding synthetic images generated by the DM, which showed a 0.02\% increase over the baseline. 
A substantial improvement was observed with the Combined DA, achieving a 3\% increase over the real baseline, with the use of images generated by the DM.

With regard to the Corn dataset (Tab.\ref{tab:TAB4}), as in the previous case, no benefit was observed from the use of Geometric DA. 
The synthetic images generated by the DM provided the best results among the synthetic-only training approaches, and their addition to the actual training set resulted in only a 1\% increase. 
Once again, the significant improvement came from the approach combining Geometric DA with synthetic additions, with DM achieving a 4\% increase over the baseline.

Finally, the PlantVillage dataset (Tab.\ref{tab:TAB4}) shows trends that differ from the previous datasets. 
In this case, although the inclusion of synthetic images provided some benefits for classification, the absolute best results were obtained through the application of Geometric DA techniques alone. 
Among DGMs, GAN demonstrated the best performance, outperforming both VAE and DM by a wide margin. 
Several factors could explain the discrepancy in these results compared to previous datasets. 
Firstly, the larger dataset size suggests that the real improvement in classification is driven by the increased diversity introduced by Geometric DA rather than the addition of synthetic images. 
Moreover, DGMs may face limitations when required to generate a large number of different classes, especially in this dataset, which includes as many as 38 classes. 
Nevertheless, it is important to highlight the excellent perceived quality of the generated images, as shown in Fig.\ref{FIG:4}.

\subsection{Impact of Increasing Synthetic Data Quantity on Classification Performance}

\begin{figure*}
	\centering
	\includegraphics[width=\columnwidth]{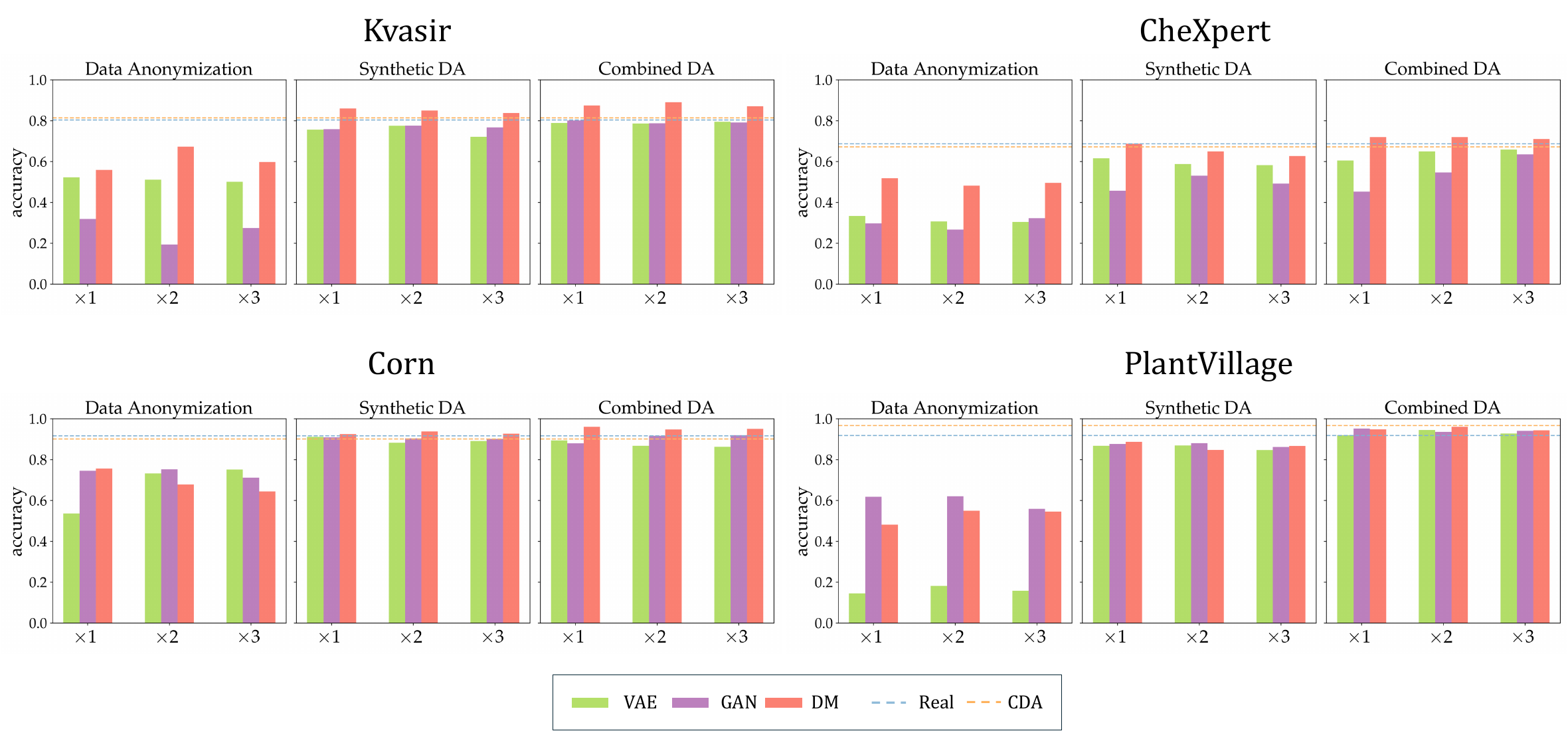}
	\caption{This figure compares the utility of synthetic data generated by the three DGMs (VAE, GAN, and DM) across the four datasets, using the accuracy of the classification task. It examines the effect of increasing the quantity of synthetic data by doubling and tripling the size of the training set. Additionally, two baselines, real data and Geometric DA, are included for reference.}
	\label{FIG:5}
\end{figure*}

Fig.~\ref{FIG:5} presents bar plots illustrating the utility results obtained from doubling and tripling the synthetic training set. 
Overall, this operation does not result in substantial differences, rather, the improvements are primarily attributed to the use of synthetic images. 

As already discussed before, while both medical datasets exhibit similar trends, classifying the 13 categories in CheXpert is considerably more difficult than classifying the categories in Kvasir.
This complexity is further emphasized by the use of top-3 accuracy for evaluating CheXpert, highlighting the difficulty of the classification task.
For Kvasir, increasing the number of images in synthetic-only training has brought clear improvements, but only for DM.
The results obtained from VAE-generated and GAN-generated images were comparable, with neither approach surpassing the actual baseline across all settings. 
However, it is noteworthy that training exclusively with VAE-generated synthetic images yielded superior performance compared to GAN-generated images, even when the number of synthetic samples was increased.
Incorporating synthetic images generated by DM into the real Kvasir training set led to substantial improvements. 
In the Synthetic DA experiment, there was a 6\% increase in accuracy over the real baseline and a 5\% increase over the Geometric DA baseline, while doubling and tripling the number of synthetic images did not yield any advantages. 
However, the best result was achieved in the Combined DA experiment, resulting in a 9\% improvement over the real baseline and an 8\% improvement over the Geometric DA baseline. 
Nevertheless, even in this case, the differences between the results obtained by varying the number of synthetic images were not substantial.

For CheXpert, increasing the number of synthetic images in the synthetic-only training did not lead to significant improvements. Similar to the previous dataset, VAE-generated images consistently outperformed GAN-generated images in terms of classification accuracy, except when training solely on a synthetic dataset containing three times the number of real images.
Synthetic DA generally resulted in lower classification performance compared to the baseline methods, except when DM-generated synthetic images were included, which led to a slight improvement. However, as in the previous case, the best performance was achieved with the Combined DA approach. Specifically, when the number of DM-generated synthetic images was twice that of the real images, the model outperformed the actual baseline and the Geometric DA baseline by 3\% and 4\%, respectively.

Similar trends to previous datasets were observed for the Corn dataset. 
In the training setting using only synthetic images for the Corn dataset, the DGMs exhibited comparable performance. 
Compared to previous datasets, however, the VAE provided better results than the GAN only in the setting where the synthetic dataset was three times larger than the real one, also outperforming the DM. 
The best result among these was achieved by the DM in the experiment with the synthetic dataset equal in size to the real dataset (0.756), which, however, is not significantly greater than the result obtained in the training with a synthetic dataset twice the size of the real one, generated by the GAN (0.753).
For the Corn dataset, there is a slight improvement in accuracy only when synthetic images generated by the DM are added, amounting to a 1\% increase over the baselines in the experiment in which the synthetic images were doubled. 
In this scenario, VAE and GAN show very similar results, in which, moreover, both do not seem to benefit from the addition of synthetic images.
As with the previous datasets, the most significant improvement is achieved by the Combined DA with images generated by DM, resulting in a 4\% increase over the baselines.

As stated above, the results for the PlantVillage dataset differ significantly from those of the previous datasets.  
However, the results provided by the variation in the number of synthetic images in the training set did not lead to differences. 
This is probably due to the fact that the numerous classes in the dataset are distinguishable in the classification without the need to add additional images, despite the limited number of samples for each class.
In training with only synthetic images, the images generated by GAN performed significantly better than VAE and DM. 
Whereas with the addition of these images to the real training set, no substantial differences were found and in no experiment was the real baseline reached. 
Finally, in all three cases, the application of Combined DA enabled the three DGMs to surpass the performance of the real baseline, though it remained below that achieved with Geometric DA.

\begin{figure*}
	\centering
	\includegraphics[width=\columnwidth]{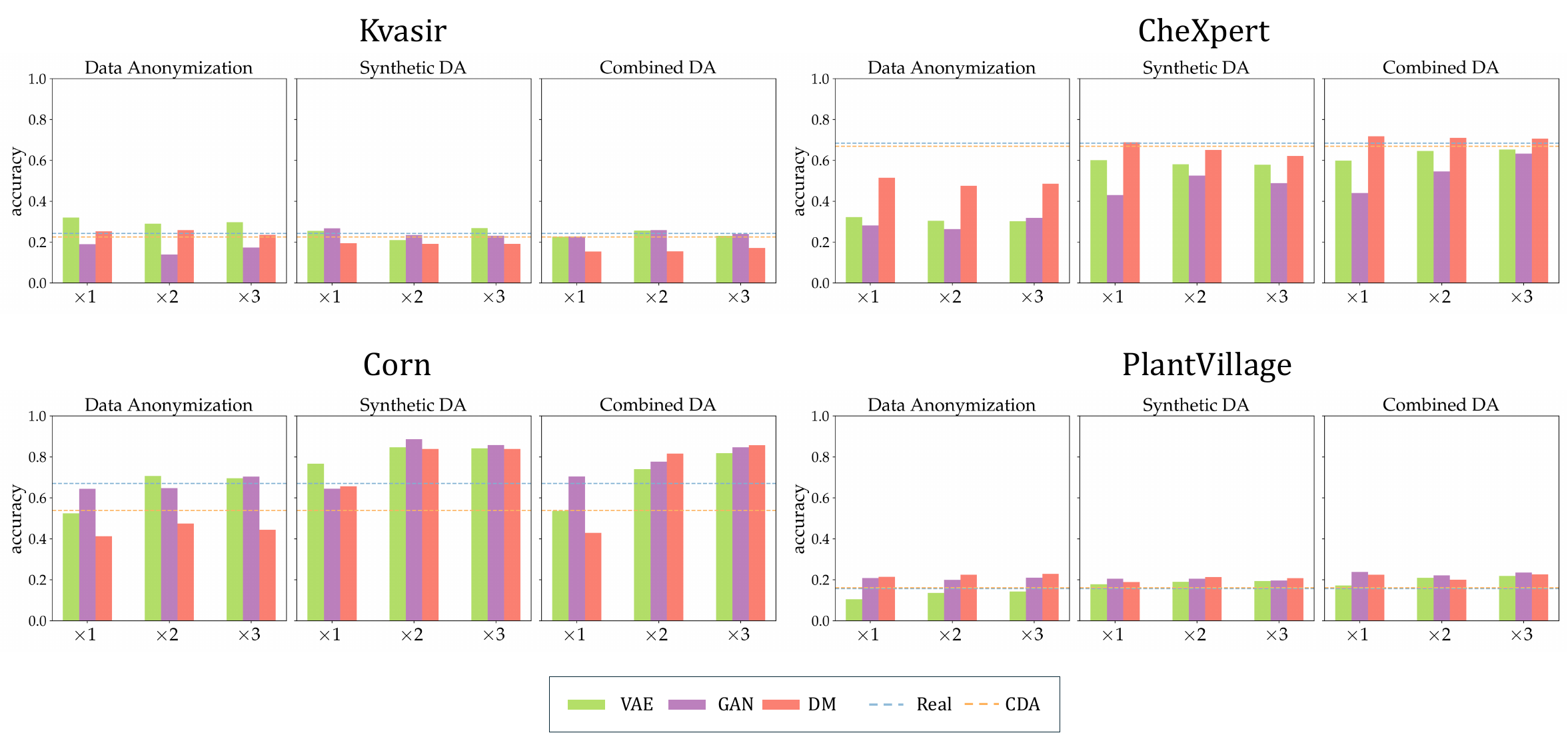}
	\caption{This figure assesses the robustness of synthetic data generated by the three DGMs (VAE, GAN, and DM) across four datasets, using adversarial accuracy determined after applying Deep Fool. It also explores the effect of increasing the amount of synthetic data by doubling and tripling the training set size. For reference, two baselines, real data and Geometric DA, are included.}
	\label{FIG:6}
\end{figure*}

From the Fig.~\ref{FIG:6} presenting the robustness results, it is evident that incorporating synthetic images during training enhances robustness across all four datasets.
However, there is no clear advantage associated with using any specific DGM over others. 
Furthermore, the results indicate that increasing the number of images in the training set, whether doubling or tripling it, does not produce significant differences in robustness. 
Instead, the enhancement primarily stems from the inclusion of synthetic images themselves, regardless of the exact quantity added.

\subsection{Key Insights and Comparative Analysis of DGMs}

\begin{figure*}
	\centering
	\includegraphics[width=\columnwidth]{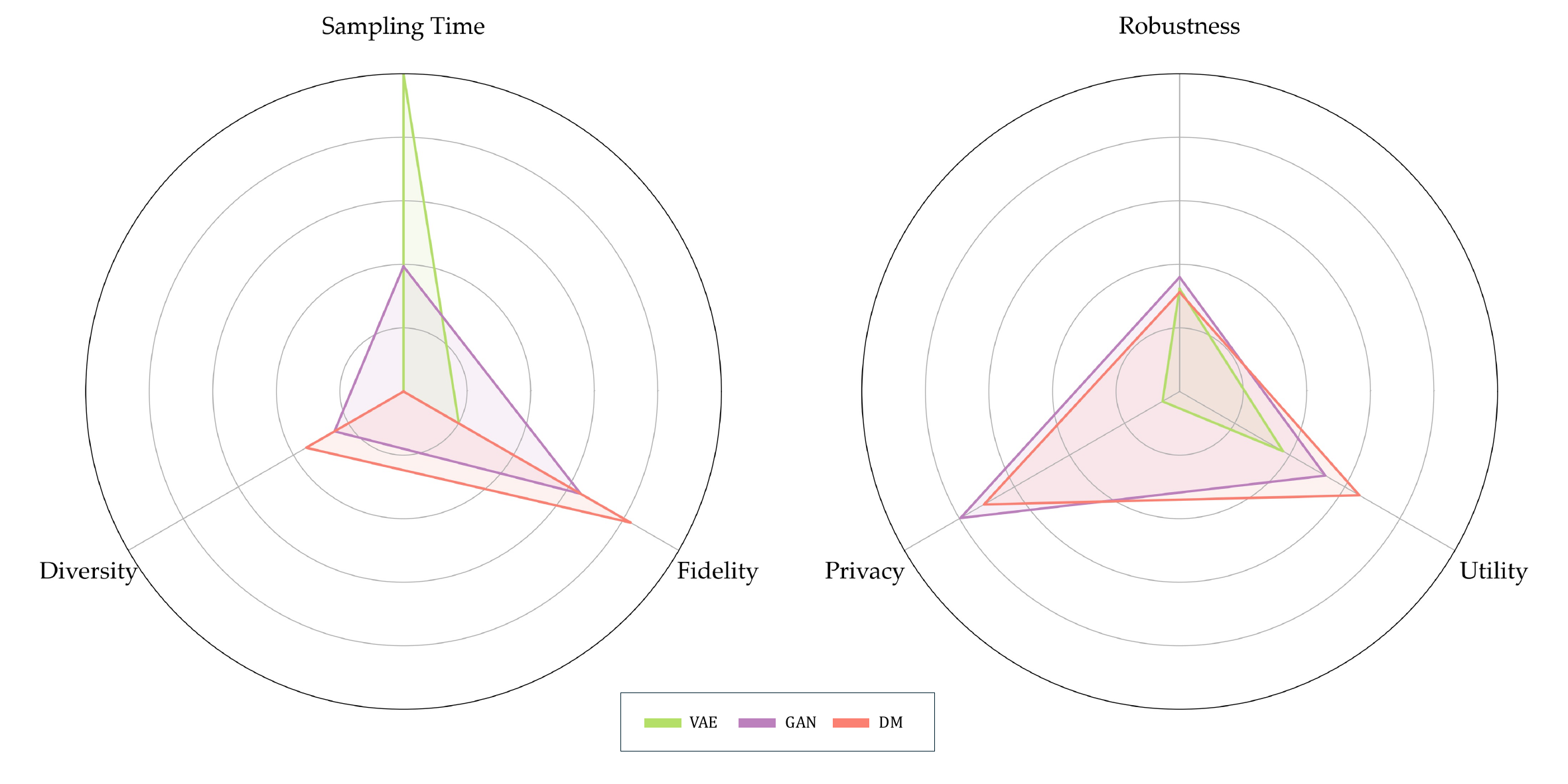}
	\caption{The radar plot provides a comparative evaluation of the three DGMs (VAE, GAN, and DM) across the six metrics. These results represent the average performance across four analyzed datasets.}
	\label{FIG:7}
\end{figure*}

The radar plot shown in Fig.~\ref{FIG:7} provides a comprehensive comparison of the performance of the three DGMs, i.e. VAE, GAN, and DM, across the six evaluation metrics. 
This visualization highlights the strengths and weaknesses of each model, offering an overall perspective on their capabilities. 
Notably, for the privacy metric, the value has been inverted to provide an immediate visual impact, as a lower value indicates better privacy preservation. 
The results depicted are averaged across four distinct datasets, providing a balanced view of how these models perform under varying conditions. 
To calculate the average value, we first computed the mean of each metric for each individual DGM across the four datasets. 
Then, we normalized each metric by taking the maximum and minimum among the 12 values for each metric (3 DGMs, 4 datasets).
In contexts of data scarcity, the generation of synthetic data becomes especially valuable. 
Overall, DGMs have succeeded in capturing the real data distribution even with limited amounts of training data, demonstrating their effectiveness in providing high-quality synthetic data that can mitigate the challenges posed by insufficient real data.

DM emerges as the top performer in terms of utility, diversity, and fidelity. 
This suggests that DM is well-suited for applications requiring high-quality and diverse synthetic data, which is especially beneficial in scenarios like medical research, where precise and varied representations are crucial for improving diagnostic models or simulating complex medical cases. 
For example, DM can be used to create synthetic MRI scans that capture a wide range of patient conditions, which aids in developing more robust diagnostic tools~\cite{dhinagar2024counterfactual}.  
Similarly, in precision agriculture, DM can generate diverse crop and soil images that capture variations in environmental conditions, which can be useful for training DL models aimed at optimizing crop yield or detecting diseases~\cite{chen2024synthetic}. 
However, it is worth noting that DM has a lower sampling speed, indicating that efficiency may be a challenge in scenarios requiring rapid data generation.

GAN, by contrast, demonstrates a balanced performance across several metrics, particularly excelling in fidelity, privacy, while maintaining a reasonable sampling speed. 
This balance makes GAN a strong candidate for medical domains, where it is crucial to generate realistic synthetic data while also ensuring privacy protection. 
Although GAN's diversity is not as high as DM, it effectively preserves privacy, making it particularly suitable for medical applications that require a focus on data confidentiality. 
The balanced performance across fidelity and privacy ensures that GAN can produce reliable representations of medical data, which can improve diagnostic and treatment models while safeguarding patient information. 
The inherent privacy benefits of GAN further make it an attractive option for healthcare scenarios, as it maintains high-quality data generation without compromising sensitive information.

VAE shows notable efficiency in sampling speed, indicating its capability for rapid data generation. 
Additionally, compared to DM and GAN, VAE has significantly fewer parameters and requires shorter training times, making it a lightweight alternative for applications with limited computational capacity. 
However, its performance falls short when it comes to fidelity and diversity, suggesting that it may not be the best option for applications where data richness and quality are crucial. 
For specific precision agriculture applications, particularly where smaller images are involved, VAE's good utility and high sampling speed make it a useful option.
For instance, in an edge computing scenario involving a drone, VAE can be utilized to rapidly generate synthetic crop images on the drone itself, which has limited computational power. 
This allows for real-time analysis and decision-making, such as identifying crop stress or detecting pest infestations, without needing to offload the data to a centralized server.

\section{Conclusions}
The purpose of this work is to underscore the importance of evaluating additional metrics for synthetic image generation beyond those commonly used in the current state of the art, while highlighting the limitations of DGMs that need to be addressed for practical, real-world applications. 
Our evaluation focuses on scenarios with limited data, using datasets from medicine and precision agriculture, fields that demand effective solutions despite data constraints. 
Our results show that each DGM has unique strengths and limitations when evaluated in terms of fidelity, diversity, sampling speed, utility, robustness, and privacy. 
The DM is particularly effective in generating synthetic images with high utility, diversity, and fidelity, making it well-suited for enhancing diagnostic models and optimizing agricultural forecasts. 
GAN balances fidelity and privacy, making it ideal for generating realistic and confidential data in sensitive medical environments. 
The VAE, while lower in fidelity and diversity, excels in sampling speed and operational efficiency, making it useful for real-time data generation in scenarios like agricultural monitoring.
The analysis insights offer practical guidance for addressing data scarcity, enabling informed decisions when selecting the most suitable model for specific tasks. 
Aligning model strengths with application needs enhances synthetic data effectiveness, emphasizing the need for continued advancements in the field.

Despite the comprehensive evaluation, this study has limitations. The analysis used a limited selection of datasets, affecting the generalizability of the findings. 
The robustness evaluation was restricted to a single type of adversarial attack, which may not fully represent potential security threats. 
Future research should expand the datasets used and include a broader array of adversarial attacks to assess model resilience comprehensively. 
Additionally, another metric could be added to evaluate privacy more effectively. 
Currently, research is predominantly focused on improving the quality of generated images. 
The primary focus of future development should be on designing new DGMs that inherently meet all six requirements simultaneously. 
Furthermore, exploring hybrid models that integrate the strengths of different DGMs could optimize performance across these metrics.
As DL evolves, updating evaluation frameworks to include new DGMs and adapting to changing regulatory and ethical standards will be essential.
Addressing these limitations and exploring future directions will enhance applicability of DGMs, improving their use in overcoming data scarcity and supporting data-driven decision-making.

\section*{Acknowledgment}
\noindent
Marco Salmè is a Ph.D. student enrolled in the National Ph.D. in Artificial Intelligence, XXXIX cycle, course on Health and life sciences, organized by Università Campus Bio-Medico di Roma.
\\
This work was partially founded by: 
i) Università Campus Bio-Medico di Roma under the program ``University Strategic Projects'' within the project ``AI-powered Digital Twin for next-generation lung cancEr cAre (IDEA)''; 
ii) PNRR MUR project PE0000013-FAIR.
iii)  Cancerforskningsfonden Norrland project MP23-1122;
iv) Kempe Foundation project JCSMK24-0094; 
\\
Resources are provided by the National Academic Infrastructure for Supercomputing in Sweden (NAISS) and the Swedish National Infrastructure for Computing (SNIC) at Alvis @ C3SE, partially funded by the Swedish Research Council through grant agreements no. 2022-06725 and no. 2018-05973.

\bibliographystyle{elsarticle-num-names} 
\bibliography{biblio}

%% The Appendices part is started with the command \appendix;
%% appendix sections are then done as normal sections
\newpage
\appendix
\section{Model Architectures for Downstream Tasks}
\label{appendixA}
To determine the most suitable architecture for the classification of the Kvasir dataset, the publication~\cite{dheir2022classification} was considered, which reports the performance of various architectures on this dataset. Among these, the one that provided the best performance was VGG16~\cite{simonyan2014very}. In this work, the network was trained from scratch to rigorously analyze the impact of synthetic data, using a batch size of 64 and a learning rate of $10^{-4}$.
For the classification of the 14 pathologies of the CheXpert dataset, a DenseNet121~\cite{huang2017densely} architecture was used. 
Irvin et al.~\cite{irvin2019chexpert} performed various experiments on CheXpert and showed that this architecture provided the best results. In this case, we use a batch size of 64, while setting the learning rate to $10^{-3}$.
In precision agriculture applications, Convolutional Neural Networks (CNN)~\cite{gu2018recent} with less deep architectures compared to the previous cases were implemented.
In the case of Corn, a sequence of 4 convolutional layers was developed, each characterized by a $3\times3$ kernel, with unit stride and padding. In the first convolutional layer there are 8 feature maps, which are then doubled in each layer. 
A ReLU activation function is used, with max-pooling layers halving the input size after each layer, and after the sequence of convolutional layers, the architecture concludes with two fully connected layers.
Regarding the classification of the PlantVillage dataset, a CNN similar to the one illustrated previously was used, but some parameters were modified considering the greater difficulty given by the classification of a much higher number of classes. The architecture is characterized by a succession of three convolutional layers, in which the number of feature maps present was doubled compared to the previous network.
For both the Corn and PlantVillage datasets, the training was conducted using a batch size of 32 and a learning rate of $10^{-3}$.
For all the classifications of each dataset, CrossEntropy was chosen as the cost function. The training of each model lasted 30 epochs, in order to ensure a consistent analysis and the best model was selected based on the lowest validation loss.

\end{document}